\documentclass[letterpaper, 10 pt, conference]{ieeeconf}  

\IEEEoverridecommandlockouts                              

\usepackage[font={footnotesize}]{caption}
\usepackage[table,dvipsnames]{xcolor}
\usepackage{multirow}
\usepackage{booktabs}

\newcommand{\ie}{i.e., }
\newcommand{\eg}{e.g., }

\usepackage{graphicx}
\usepackage{hyperref}
\usepackage{adjustbox}
\usepackage{threeparttable}
\usepackage{blindtext}
\usepackage[absolute]{textpos}
\usepackage{amssymb}
\usepackage{subcaption}

\usepackage{xcolor}
\hypersetup{
    colorlinks,
    linkcolor={red!70!black},
    citecolor={blue!50!black},
    urlcolor={blue!70!black}
}

\graphicspath{{./images/}}
\DeclareGraphicsExtensions{.pdf,.jpeg,.png}

\title{\LARGE \bf
Data Limitations for Modeling Top-Down Effects on Drivers' Attention
}

\author{Iuliia Kotseruba and John K. Tsotsos
\thanks{Both authors are with the EECS Department of Computer Science and Electrical Engineering, York University, Canada.
        Email: {\tt \{yulia84, tsotsos\}@yorku.ca}}%
}

\begin{document}

\begin{textblock}{10}(1,0.5)
\noindent \footnotesize Accepted at IEEE Intelligent Vehicles Symposium (IV), 2024
\end{textblock}

\maketitle
\thispagestyle{empty}
\pagestyle{empty}

\begin{abstract}
Driving is a visuomotor task, i.e., there is a connection between what drivers see and what they do. While some models of drivers' gaze account for top-down effects of drivers' actions, the majority learn only bottom-up correlations between human gaze and driving footage. The crux of the problem is lack of public data with annotations that could be used to train top-down models and evaluate how well models of any kind capture effects of task on attention. As a result, top-down models are trained and evaluated on private data and public benchmarks measure only the overall fit to human data.

In this paper, we focus on data limitations by examining four large-scale public datasets, DR(eye)VE, BDD-A, MAAD, and LBW, used to train and evaluate algorithms for drivers' gaze prediction. We define a set of driving tasks (lateral and longitudinal maneuvers) and context elements (intersections and right-of-way) known to affect drivers' attention, augment the datasets with annotations based on the said definitions, and analyze the characteristics of data recording and processing pipelines w.r.t. capturing what the drivers see and do. In sum, the contributions of this work are: 1) quantifying biases of the public datasets, 2) examining performance of the SOTA bottom-up models on subsets of the data involving non-trivial drivers' actions, 3) linking shortcomings of the bottom-up models to data limitations, and 4) recommendations for future data collection and processing. The new annotations and code for reproducing the results is available at \url{https://github.com/ykotseruba/SCOUT}.
\end{abstract}

\section{Introduction}

Driving is a visuomotor task where perception, especially vision \cite{sivak1996information}, and motor actions are tightly intertwined \cite{cunningham2001driving, lappi2017systematic, lappi2018visuomotor}. Despite ample experimental evidence that driving task and context affect how drivers observe the traffic scene \cite{shinoda2001controls,baluch2011mechanisms,metz2017driving}, many of the state-of-the-art (SOTA) models are bottom-up \cite{kotseruba2022practical}. In other words, they establish a mapping between the driving footage and human ground truth without considering the underlying decision-making processes and motor actions performed by the drivers \cite{ning2019efficient, 2020_T-ITS_Deng, 2018_PAMI_Palazzi, 2021_T-ITS_Fang}. Whether these processes are captured implicitly cannot be assessed due to lack of appropriate annotations in public benchmarks. Such annotations could also be useful for explicit modeling of drivers' actions and scene attributes. But since none are available, existing top-down models \cite{2021_T-ITS_Amadori, 2021_ICCV_Baee} are developed using private datasets with additional information or unpublished labels for public data.

Due to the importance of data in model development and evaluation, this paper focuses on the issues of annotations for top-down influences on drivers' attention and analysis of the contents and quality of public datasets w.r.t to task and context. Our previous work \cite{kotseruba2023understanding} addressed some of these problems by correcting and extending annotations for the DR(eye)VE dataset \cite{2018_PAMI_Palazzi}. Here, we extend the analysis and apply it to three more public datasets. 

In sum, we delve deeper into the properties of the recorded data w.r.t. capturing what drivers perceived, what actions they performed, and context in which these actions were made. Specifically, we identify the limitations of recording equipment, recording conditions, eye-tracking data processing pipelines, and annotation procedures. Since only the latter can be remedied to some extent without collecting new data, we propose a new set of annotations for drivers' actions and traffic context. Using these new annotations, we demonstrate that most available datasets are dominated by simple scenarios where the vehicle maintains speed/lane or is stopped. We then show that the bottom-up models for drivers' gaze prediction do not accurately capture effects of drivers' actions and context, and link some aspects of model performance to the limitations of the data. We conclude with recommendations for future data collection efforts.

\section{Related work}
\noindent
\textbf{Task and context definition.} We begin by defining task and context elements. Drivers' \textit{task} is an action or a sequence of actions that lead to maintaining speed/lane, remaining stationary, moving the vehicle laterally (turns or lane changes), or longitudinally (accelerating and decelerating). \textit{Context} refers to the physical and temporal scenario for the task. Here, we focus on two aspects of context: intersection type (signalized, unsignalized, highway ramp) and driver’s priority w.r.t. other road users (right-of-way, yield). Both have been shown to affect drivers' gaze distribution \cite{2014_JEMR_Lemonnier,2015_TR_Lemonnier}. These definitions will be used for the rest of the paper.

\noindent
\textbf{Datasets for drivers' gaze prediction.} Large-scale public datasets stimulated innovation in many areas of artificial intelligence and computer vision, and driving attention research is no exception. Based on our estimates, there are over forty public datasets \cite{attention_driving_datasets} that capture aspects of drivers' attention. For this paper, we focus on a subset of datasets that contain driving footage and human eye-tracking data. Specifically, we analyze DR(eye)VE \cite{2018_PAMI_Palazzi}, BDD-A \cite{2018_ACCV_Xia}, MAAD \cite{2021_ICCVW_Gopinath}, and LBW \cite{2022_ECCV_Kasahara}\footnote{One other dataset, DADA-2000 \cite{2021_T-ITS_Fang}, is hosted on Baidu Wangpan which is not accessible to the authors.}. All datasets are comparable in volume (4--6 hours of video footage and corresponding eye-tracking data) and are recorded in North America and Europe.

These datasets represent different approaches to gathering data. DR(eye)VE and LBW are collected in on-road conditions while the drivers whose gaze was recorded controlled the vehicle. BDD-A and MAAD are collected in the lab while human subjects passively viewed driving footage recorded with dashboard cameras on the computer screen. The datasets feature diverse locations, visibility, and weather conditions. DR(eye)VE and LBW consist of long clips recorded in low- to mid-density traffic, BDD-A contains short clips sampled around evasive or braking maneuvers in Berkeley DeepDrive Video (BDDV) dataset \cite{xu2017end} and features low to dense traffic.

\noindent
\textbf{Additional annotations.} DR(eye)VE and BDD-A, as some of the early datasets for studying drivers' attention, have been used to develop and evaluate dozens of models. However, neither these datasets nor the more recent MAAD and LBW contain fine-grained annotations relevant to the driving task and context. Only DR(eye)VE offers textual annotations: video-level labels for weather, time of day, and location. In addition, some frames where saliency maps deviate from the average gaze distribution (approx. 20\% of the data) are marked as \textit{recording errors}, \textit{inattentive}, \textit{subjective}, and \textit{acting}. The latter category refers to some unspecified actions performed by the driver. Similarly, in BDD-A, frames deviating from the average are given higher weight during training since they are associated with drivers' responses to traffic signals or actions of other road users.  However, neither the implementation of per-frame weights nor corresponding labels are available.

\noindent
\textbf{Evaluation of top-down effects.} Several past works estimated the impact of different drivers' actions and scenarios on model performance. For example, \textit{acting} annotations in DR(eye)VE mentioned earlier have been used for evaluation \cite{2018_PAMI_Palazzi, 2018_ACCV_Xia}. Other approaches to partitioning the data have also been attempted. In \cite{2021_ICCV_Baee}, test sets of DR(eye)VE and BDD-A were further split into categories corresponding to three tasks---merging-in, lane-keeping, and braking. The authors of \cite{2018_ITSC_Tawari} used yaw rate (derived from GPS) as a proxy for lateral actions in DR(eye)VE. However, neither of the works makes the labels or methods for generating them available.

\section{Characteristics of the datasets}
\label{sec:data_analysis}

In this section, we analyze the recording and processing pipelines of DR(eye)VE \cite{2018_PAMI_Palazzi}, BDD-A \cite{2018_ACCV_Xia}, MAAD \cite{2021_ICCVW_Gopinath}, and LBW \cite{2022_ECCV_Kasahara} with respect to capturing driving task and context defined earlier. In Sections \ref{sec:video_data}, \ref{sec:vehicle_data}, and \ref{sec:data_recording} we focus on the common shortcomings of the recorded video, vehicle telemetry, and eye-tracking data, respectively. Section \ref{sec:data_processing} highlights issues with processing and aggregation of eye-tracking data for training and evaluation. Table \ref{tab:dataset_properties} summarizes data properties, which are discussed in more detail below.

\begin{table}[]
\centering
\caption{Properties of the selected datasets that contain driving footage and eye-tracking data.}
\label{tab:dataset_properties}
\begin{adjustbox}{max width=\columnwidth}
\begin{threeparttable}
\begin{tabular}{@{}clcccc@{}}
\toprule
\multicolumn{2}{c}{}     & \textbf{DR(eye)VE} \cite{2018_PAMI_Palazzi} & \textbf{BDD-A} \cite{2018_ACCV_Xia} & \textbf{MAAD}\textsuperscript{a} \cite{2021_ICCVW_Gopinath} & \textbf{LBW} \cite{2022_ECCV_Kasahara}  \\ \midrule
\multirow{9}{*}{\begin{tabular}[c]{@{}c@{}}Video \\ data\end{tabular}}       & \# videos      & 74                 & 1435           & 106           & 74            \\
                             & \# frames          & 555K      & 405K            & 794K            & 122K             \\
                             & Continuous         & yes       & no              & yes             & no               \\
                             & Video len (s)   & 300       & 11 (2)          & 299 (5)         & 329 (76)         \\
                             & Segment len (s) & 300       & 11 (2)          & 299 (5)         & 2 (3)            \\
                             & Scene view         & yes       & yes             & yes             & yes              \\
                             & Driver’s view      & yes       & no              & no              & no               \\
                             & In-cabin view      & no        & no              & no              & yes              \\
                             & Traffic			  & light-med & light-heavy     & light-med       & light            \\ \midrule
\multirow{3}{*}{\begin{tabular}[c]{@{}c@{}}Scene\\ camera\end{tabular}}      & Image size     & 1920$\times$1080          & 1280$\times$720       & 1920$\times$1080     & 942$\times$489       \\
                             & Hz                 & 25        & 30              & 25              & 5                \\
                                                                             & Field of view  & \textless 120      & \textless 120  & \textless 120 & \textless 120 \\ \midrule
\multirow{7}{*}{\begin{tabular}[c]{@{}c@{}}Eye-\\ tracking\\ data\end{tabular}} & Where recorded & on-road            & in-lab         & in-lab        & on-road       \\
                             & Hz                 & 60        & 1000            & 1000            & 5                \\
                             & \# subj. per video & 1         & $\geq$4              & 1 – 11          & 1                \\
                             & \# of subjects     & 8         & 45              & 23              & 28               \\
                             & Subjects’ age      & 20–40     & -               & 20-55           & \textgreater{}18 \\
                             & Subjects’ gender   & 7M, 1F    & -               & 22M, 6F         & -                \\
                             & Licensure (years)  & -         & \textgreater{}1 & \textgreater{}2 & -                \\ \midrule
\multirow{3}{*}{\begin{tabular}[c]{@{}c@{}}Vehicle\\ data\end{tabular}}      & GPS (Hz)       & 1                  & 1              & 1             & -             \\
                             & Speed (Hz)         & 25        & 1               & 25              & -                \\
                             & Heading (Hz)       & 25        & 1               & 25              & -                \\ \midrule
\multirow{2}{*}{Labels} & Action             & partial   & no              & no              & no               \\
                             & Video attributes   & yes       & no              & yes             & no               \\ \bottomrule
\end{tabular}%
\begin{tablenotes}
\footnotesize
\item[a] MAAD dataset is composed of 74 original DR(eye)VE scene videos and their copies with various manipulations applied (e.g. blurred, rotated). Eye-tracking data was recorded as subjects viewed the videos on the monitor.
\end{tablenotes}
\end{threeparttable}
\end{adjustbox}
\end{table}

\subsection{Video data collection}
\label{sec:video_data}
Video footage should capture what the driver sees, ideally, as closely as possible. Therefore, the following properties of video data are most relevant: spatial context (camera angle of view and position), temporal context (duration and continuity), and video quality.

\begin{figure}
\centering
\includegraphics[width=\columnwidth]{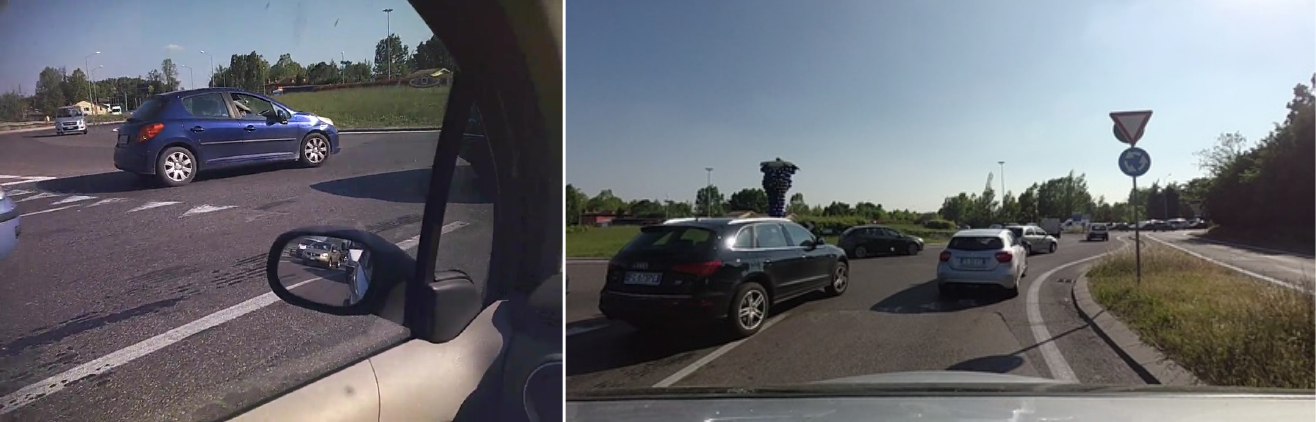}
\caption{An example from DR(eye)VE, showing the view of the roundabout from the camera mounted on the driver's head (left) and from the vehicle forward-facing camera (right). Note that the forward camera does not capture what the driver can see from the side window by turning their head.}
\label{fig:driver_view}
\vspace{-1.5em}
\end{figure}

\noindent
\textbf{Camera limitations in capturing drivers' view.} Driving footage in all datasets is recorded with monocular dashboard cameras, except LBW, which is captured with a stereo pair of Azure Kinect RGB-D cameras (only one camera's output is made available). All cameras used have at most 120$^\circ$ horizontal field of view, which falls short of the human binocular visual field spanning 200$^\circ$ horizontally and 125$^\circ$ vertically \cite{2020_iPerception_Strasburger}. Therefore, only a portion of what the driver perceives (even without head movements) can be recorded. Consequently, this may be detrimental to data analysis and model training. For example, drivers can turn their head to observe look out of the side windows, but these areas and any events that attracted drivers' attention will be absent in the recording (see Figure \ref{fig:driver_view}). The same applies to the view behind the vehicle in the side and rearview mirrors.

\begin{figure}
\centering
\includegraphics[width=0.493\columnwidth]{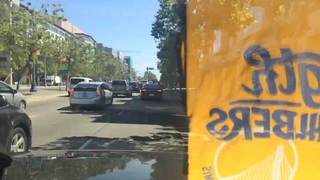}
\includegraphics[width=0.493\columnwidth]{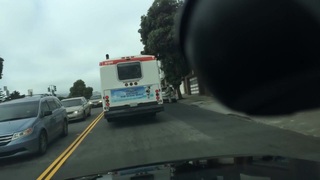}
\caption{BDD-A videos with quality issues: reflection on the windshield (left) and tilted and obstructed view (right).}
\label{fig:bdda_issues}
\vspace{-1em}
\end{figure}

Camera location differs across datasets. In DR(eye)VE, a camera is installed on the rooftop of the vehicle. Due to being at a higher vantage point, the recording often does not match what the driver was seeing, which causes issues with data processing (see Section \ref{sec:data_processing}). In LBW and BDD-A, the camera is installed inside the car above the dashboard and better approximates drivers' view. While in LBW and DR(eye)VE, the camera position is fixed, in BDD-A, positioning of the camera varies widely across videos, with more than 150 videos (10\% of the dataset) recorded with the camera off-center, tilted, or partially obstructed (see examples in Figure \ref{fig:bdda_issues}). 

\noindent
\textbf{Limited temporal context.} Only DR(eye)VE contains continuous recordings spanning 5 minutes each. This duration allows capturing drivers' quick reactions and deliberate actions, as well as events preceding and following them. 

LBW videos likewise span $\approx5$ minutes, but recordings are not continuous. Using the indices of the first and last frame of the videos to infer their duration, we estimated that 26.5\% of the frames are missing, fracturing 74 videos into 11K segments of 2s (10 frames). Crucially, data loss is distributed unequally: 16--19\% of frames are missing from episodes with longitudinal actions and 32--35\% during lateral maneuvers (based on the annotations from Section \ref{sec:action_labels}).

In contrast, BDD-A consists of 10s clips extracted from the larger BDDV dataset around braking events. Because of this design decision, many videos end abruptly, \eg before the intersection or in the middle of the turn. Furthermore, multiple videos contain repeated or dropped frames.

\begin{figure}
\centering
\includegraphics[width=0.491\columnwidth]{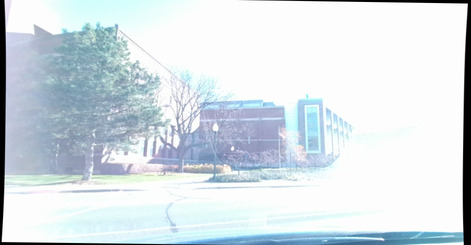}
\includegraphics[width=0.491\columnwidth]{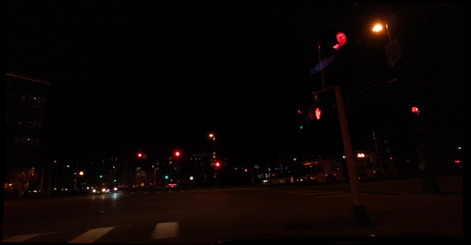}
\caption{Overexposed (left) and underexposed (right) frames in LBW.}
\label{fig:lbw_issues}
\vspace{-1em}
\end{figure}

\noindent
\textbf{Uneven video quality.} Lastly, video quality varies across datasets. In BDD-A, which is based on recordings crowdsourced from hundreds of drivers, nearly 37\% of all videos have issues, such as blurriness, reflections, obstructions, and compression artifacts (see Figure \ref{fig:bdda_issues}). The videos in DR(eye)VE and LBW are recorded in more controlled environments and are thus more consistent. However, in LBW, about 6K frames (5\% of all data) are overexposed and two nighttime videos (comprising $\approx$2K frames) are underexposed, making it difficult to discern road markings, signs, signals, and vehicles (see Figure \ref{fig:lbw_issues}). In DR(eye)VE, the placement of the camera outside the vehicle is problematic for rainy scenarios since water drops accumulate on the camera lens and obstruct the view.

\subsection{Vehicle data collection} 
\label{sec:vehicle_data}

Vehicle telemetry is a direct reflection of drivers' intentions and actions. Current speed, status of the brake or acceleration pedals, the steering wheel rotation, IMU and GPS signals are commonly used in the literature as a proxy of what the driver is doing, whereas turn signals indicate drivers' intention to change lane or make a turn. 

\noindent
\textbf{Minimal, coarsely sampled, and noisy vehicle data.} DR(eye)VE and BDD-A provide only a small set of vehicle data, namely, GPS coordinates, heading, and speed, most of which are coarsely sampled at 1Hz (see Table \ref{tab:dataset_properties}). While this data can be used to infer general route information and longitudinal actions, detection of lateral maneuvers, such as lane changes, requires an IMU signal.

Moreover, some vehicle data is either missing or noisy. In BDD-A, it is absent for 90 videos (approx. half of the validation set). In several of the DR(eye)VE videos, vehicle data lags noticeably (in one instance, by nearly a minute). Lastly, LBW does not provide any vehicle information. 

\subsection{Recording eye-tracking data} 
\label{sec:data_recording}

Eye-tracking data represents (with caveats \cite{kotseruba2021behavioral}) what the driver looked at, and can shed light on the unobservable decision-making processes. However, there is a trade-off between quality and ecological validity of eye-tracking data acquisition, depending on the recording equipment (mobile or static eye-tracker) and conditions (on-road or in-lab). 

\textbf{Sparsity of on-road gaze recordings.} On-road setup is the most ecologically valid, since the driver controls the car and their decisions have consequences. At the same time, rides cannot be replicated across multiple participants and only mobile or remote eye-trackers can be used to allow for free head and body movement. As a result, recorded gaze data is sparse and has lower precision and sampling rate than what can be achieved in the lab. For example, gaze data for DR(eye)VE and LBW recorded in actual vehicles on the road is sampled at $50$Hz, resulting in few data points per frame. In LBW, it is further downsampled to 5Hz and a single gaze coordinate per frame.

\textbf{Lack of engagement and limited context in the laboratory conditions.} In the lab, it is possible to recreate the same conditions across participants and use more restrictive but accurate recording equipment. For example, to record ground truth for  BDD-A and MAAD, multiple subjects passively viewed the same videos on the computer monitor while their gaze was registered with static eye-trackers at 1000Hz. 

\begin{figure}
\centering
\includegraphics[width=\columnwidth]{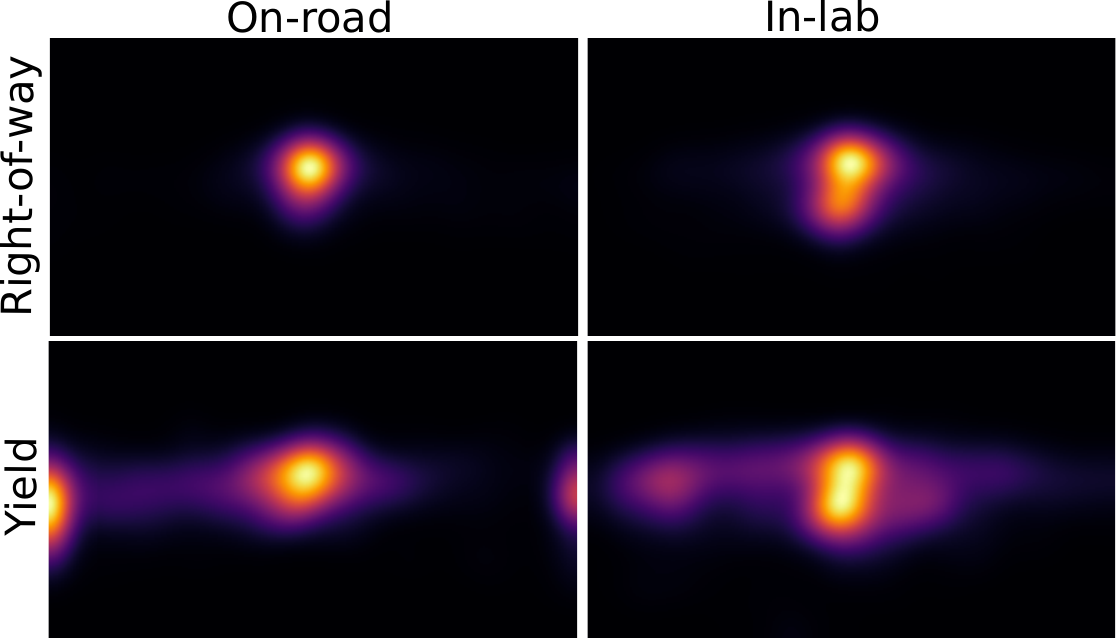}
\vspace{-1em}
\caption{Gaze recorded on-road (from DR(eye)VE) and in-lab (from MAAD) for right-of-way and yielding episodes.}
\label{fig:road_lab}
\vspace{-1em}
\end{figure}

\begin{figure}
\centering
\includegraphics[width=\columnwidth]{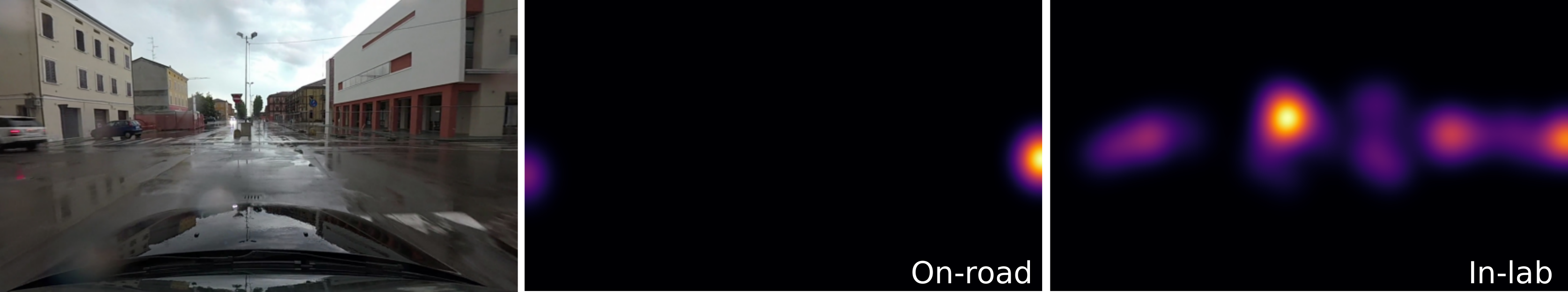}
\vspace{-1em}
\caption{Example of gaze recorded on-road and in-lab aggregated over one yielding scenario.}
\label{fig:road_lab_sample}
\vspace{-1.5em}
\end{figure}

The in-lab setup has many known limitations \cite{ho2014extent}. Here, the main ones are lack of full context and engagement in the task, which impact how human subjects observe the scene. We will demonstrate this using DR(eye)VE and MAAD, that provide eye-tracking data recorded on-road and in-lab, respectively, for the same videos. Figure \ref{fig:road_lab} shows on-road and in-lab fixations aggregated across all scenarios where driver had right-of-way or yielded. Note that right-of-way scenarios look similar, but on yielding episodes there are differences. Specifically, in data recorded on the road, there are peaks at either side of the image frame, corresponding to driver's gaze falling on side mirrors or side windows. In comparison, subjects in the lab only have access to what the scene camera has recorded. Since they cannot see beyond the image frame, their viewing patterns are more center-biased. Figure \ref{fig:road_lab_sample} shows a three-way junction, where driver is repeatedly checking the intersecting road, whereas in-lab subjects focus more on the center of the frame or passing vehicles. 

Similar differences in gaze patterns are observed for lane changes, where recordings do not show parts of the scene visible to the driver through side windows and mirrors.

\subsection{Processing eye-tracking data}
\label{sec:data_processing}

For the gaze prediction task, raw eye-tracking data is processed and aggregated into per-frame heatmaps (or saliency maps) representing areas where the human subjects fixated. However, in each dataset, it is done differently.

\noindent
\textbf{Noisy gaze transformation from the head-mounted eye-tracker view to scene image.} In DR(eye)VE and LBW, which use mobile eye-trackers, gaze is recorded with respect to the driver's frame of reference (D). The camera installed on the vehicle offers a more stable view, therefore gaze from the driver's perspective is transferred to the scene view (S) via a homography (since the cameras are not calibrated). 

\begin{figure}
\centering
\includegraphics[width=\columnwidth]{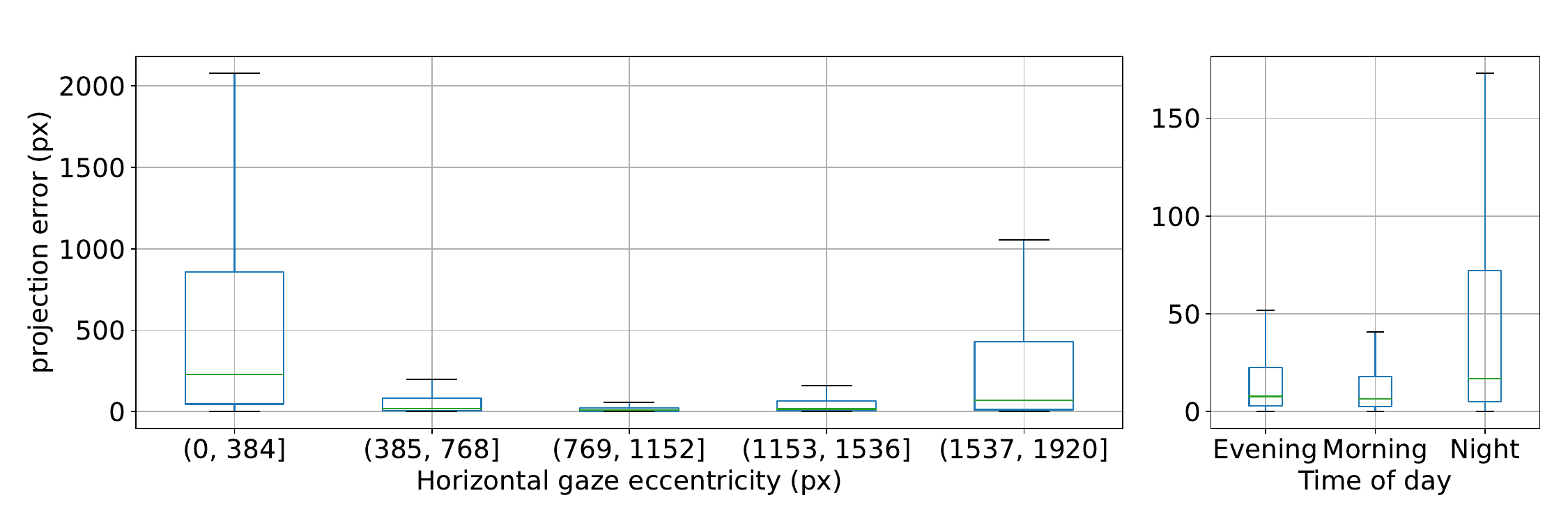}
\vspace{-1.5em}
\caption{DR(eye)VE projection errors from the drivers' to scene view for different horizontal gaze eccentricity (left) and time of day (right). Outliers are not plotted for clarity.}
\label{fig:etg2gar_boxplots}
\vspace{-1em}
\end{figure}

\begin{figure}
\centering
\includegraphics[width=\columnwidth]{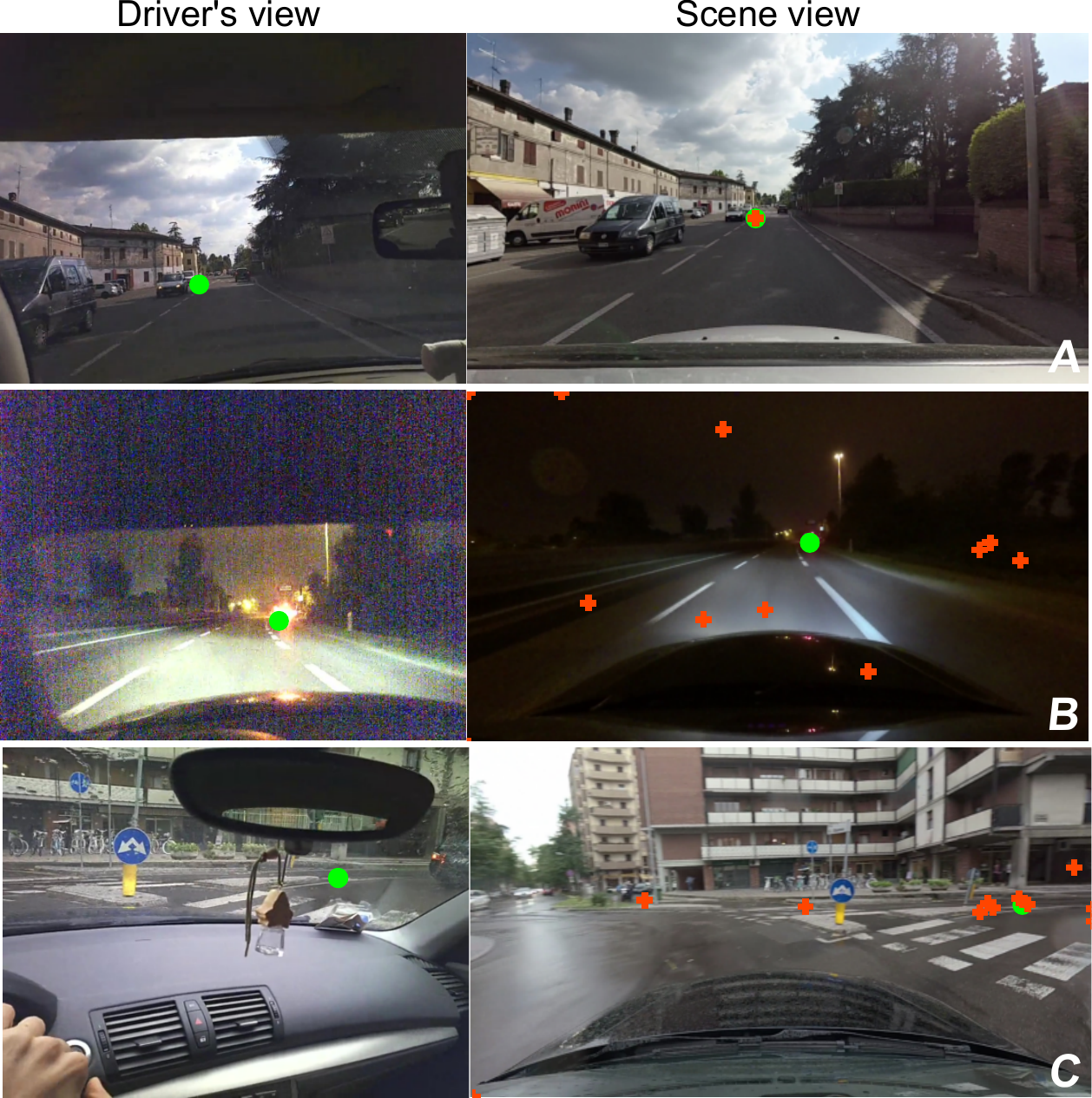}
\vspace{-1em}
\caption{Gaze projection errors in DR(eye)VE. Errors are low when projections (red crosses) are close to true gaze location (green circles) in the scene view. Note that errors increase from day (A) to night (B) scene, even in the simple case when the driver maintains speed and looks straight ahead. During turns (C), matches are noisy due to the driver's head rotation, which reduces the overlap between two views.}
\label{fig:etg2gar_vis}
\vspace{-1.5em}
\end{figure}

The homography calculation is inherently noisy since it depends on feature detection and selection. In \cite{2018_PAMI_Palazzi} the theoretical upper bound for S-D homography transformation is estimated as 200 px (for 1920×1080 image). Empirically, we determined that median projection error for most videos is within 20 px\footnote{To measure the projection error, we used data from DR(eye)VE that was temporally re-aligned, cleaned up, and manually corrected as described in \cite{kotseruba2023understanding}. We then selected from each video 1000 random pairs of driver and scene frames with at least one valid fixation (74,000 frames in total or 13\% of the data), ran a homography transformation 10 times on each pair of frames, and measured the deviation from the ground truth fixation as Euclidean distance in pixels.}, but over 13\% of the frames contain outliers, half of which exceed 200 px, particularly in night and/or rainy conditions. Errors also increase with horizontal gaze eccentricity (Figure \ref{fig:etg2gar_boxplots}), \ie the fixations closer to the edges of the scene are more difficult to map or even fall outside the camera viewpoint. Such cases usually occur when the drivers rotate their head, which may cause motion blur and reduced overlap between the drivers' and scene views (Figure \ref{fig:etg2gar_vis}).

\begin{figure}
\centering
\includegraphics[height=2.8cm]{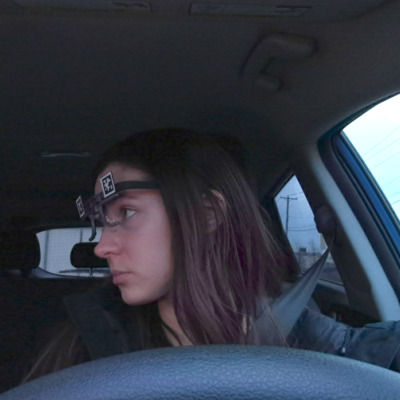}
\includegraphics[height=2.8cm]{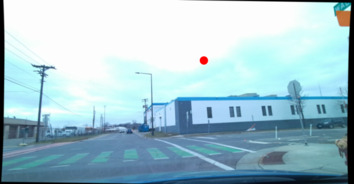}
\caption{Frame from the LBW dataset with incorrectly projected driver's gaze. Since the video from eye-tracking glasses is not available, we show a frame from the driver-facing camera (left) and a corresponding scene view (right) with the driver's gaze location shown as a red circle.}
\label{fig:lbw_proj_error}
\vspace{-1.5em}
\end{figure}

In LBW, the same approach is used to project the drivers' gaze from the eye-tracking glasses onto the scene. While gaze is projected incorrectly in many frames with large head rotations (\eg Figure \ref{fig:lbw_proj_error}), the exact errors could not be estimated since neither the raw eye-tracker output nor the driver's eye-tracking camera view is available.

\noindent
\textbf{Inclusion of blinks, saccades, and gaze towards vehicle interior.} While blinks and saccades do not have direct bearing on the drivers' task, they still make up a significant proportion of the data---27\% in DR(eye)VE and 18\% in MAAD. Since vision is suppressed during blinks and saccades, not removing them may increase noise. 

Fixations towards the interior of the vehicle also contribute to noise when projected onto the scene. This is not a concern for BDD-A and MAAD since the videos do not show much of the cars' interior. But in DR(eye)VE, 3\% of all fixations are directed towards the vehicle's cabin \cite{kotseruba2023understanding}, varying from 0\% to 10\% per video, depending on the drivers' individual preferences and the environment. For LBW, it is not reported whether in-vehicle fixations were excluded.

At the same time, in-vehicle gaze may also contain useful information. Drivers' glances towards elements of the vehicle interior, such as dashboard, mirrors, and infotainment screen, can be used to infer their intentions and assess their situation awareness. For example, lateral maneuvers, such as lane changes and turns, must be preceded by checking the rearview and side mirrors. 

\noindent
\textbf{Differences in ground truth saliency maps across datasets.} The final step is combining the fixations into saliency maps. In the saliency literature, a typical processing contains the following steps: non-saccadic eye movements (fixations, pursuit) of multiple subjects observing the same stimuli are filtered (to remove outliers), aggregated per image or frame (if stimuli were videos), and convolved with a Gaussian filter with kernel roughly equal to the size of the fovea in the given setup \cite{mit-saliency-benchmark, mathe2014actions, xu2014predicting, nguyen2019saliency}.

\begin{figure}
\centering
\includegraphics[width=\columnwidth]{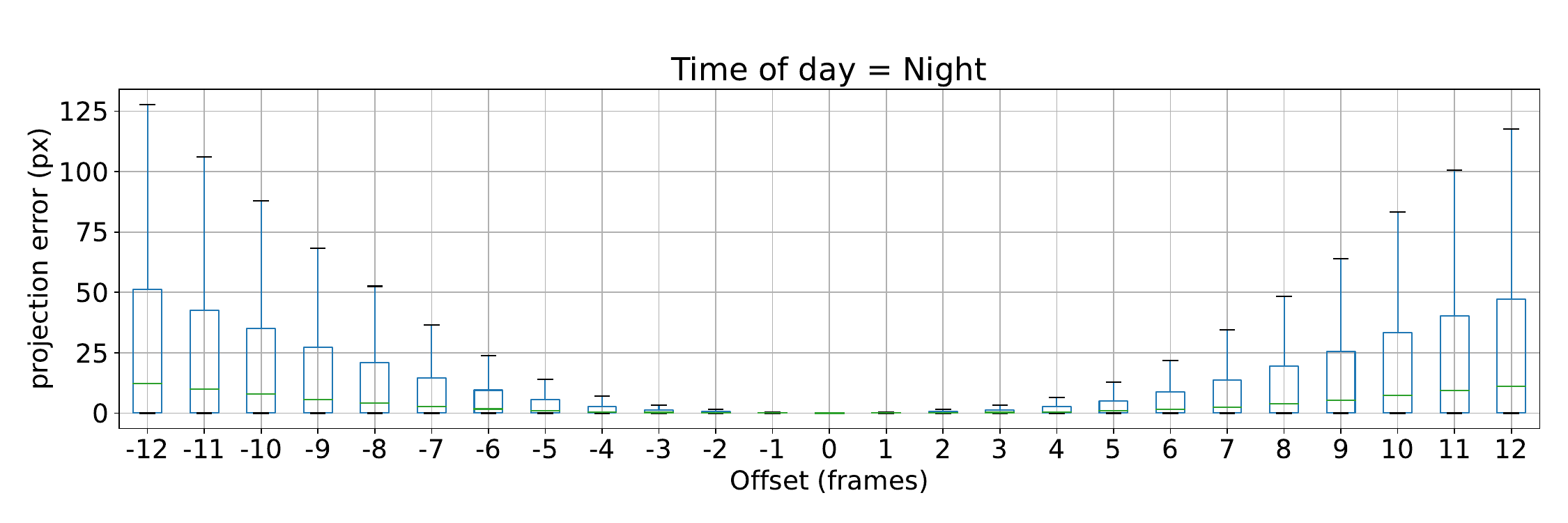}
\vspace{-1em}
\caption{Homography projection errors in DR(eye)VE night videos within the 25 frame (1s) temporal window, centered on the 0-th key frame. Outliers are not plotted for clarity.}
\label{fig:gar2gar_boxplots}
\vspace{-1em}
\end{figure}

\subsubsection{BDD-A}
BDD-A is the most similar to video saliency datasets and follows the same procedure for generating the maps, with one change: in addition to  spatial dimension, Gaussian smoothing is also applied temporally (with kernels set to 25 px and 4 frames, respectively).

\subsubsection{DR(eye)VE} Since DR(eye)VE is recorded on the road, each video has only a single observer. Consequently, there are at most few fixations per frame or none at all (if the driver blinked or moved their eyes), which in turn results in sparse saliency maps, if the usual methods are applied. 

In DR(eye)VE, this problem is resolved by aggregating fixations within a temporal window of 1s (25 frames), i.e., for each frame $t$, the saliency map contains fixations from 12 preceding and 12 following frames. Since the camera may move, a homography computation is applied to map all fixations within the temporal window $[t-12, t+12]$ to the key frame $t$. However, even within a short time window, changes in the scene can still be significant due to interactions between moving objects and the camera ego-motion\footnote{We estimated median error for the temporal homography similarly to S-D. Since manual ground truth is prohibitively labor-intensive to generate for each frame, we instead compute the mean coordinate of all 10 projected points as a reference.}. As a result, the median projection error increases towards the edges of the temporal window, but does not exceed 10 px. At the same time, 12--14\% of outlier errors exceed 100 px. In the night videos, the median error increases and although the percentage of outliers remains the same, most are \textgreater 200 px (see Figure \ref{fig:gar2gar_boxplots}). Once the fixations are aggregated across the temporal window for every frame, a max over Gaussian response function is computed. 

\subsubsection{LBW} In LBW, which is also recorded on the road, the data is even sparser since both videos and eye-tracking data are subsampled to 5Hz. Thus, each saliency map is based on the 3D coordinates of a \textit{single} fixation combined with the depth map of the scene. Due to the use of a large smoothing kernel, ground truth covers some of the peripheral field of view, different from BDD-A and DR(eye)VE that aim to represent foveal vision.

\begin{figure}
\includegraphics[width=\columnwidth]{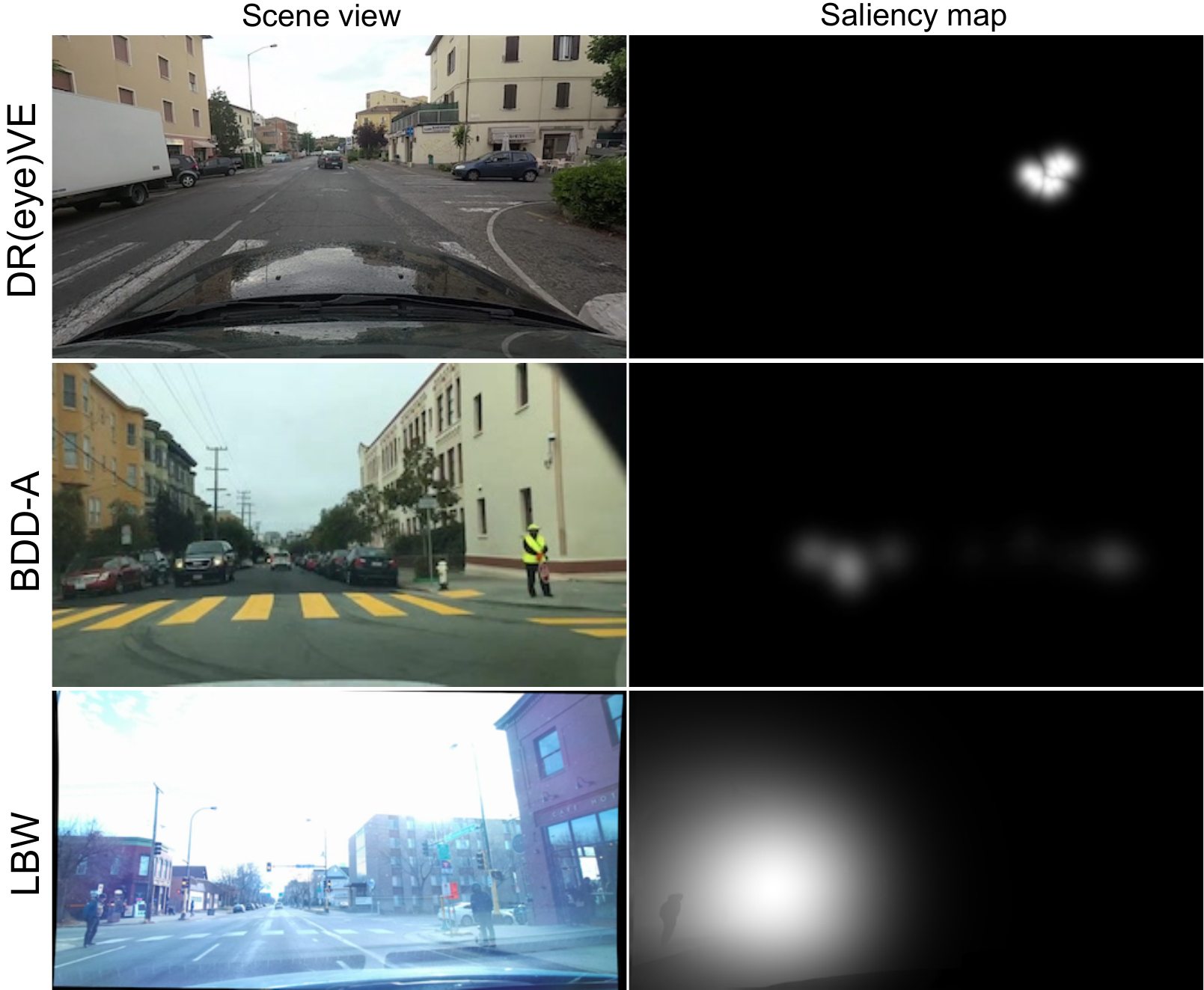}
\caption{Scene images and ground truth saliency maps.}
\label{fig:salmap_examples}
\vspace{-1.5em}
\end{figure}

These differences are further illustrated in Figure \ref{fig:salmap_examples}, which shows examples of the frames and corresponding ground truth saliency maps for each dataset.

\section{Contents of the datasets}

\label{sec:extended_annotations}

In this section, we discuss how definitions for task and context introduced in Section \ref{sec:data_analysis} were used to annotate DR(eye)VE, BDD-A, and LBW, and analyze their contents.  

\subsection{Driving task}
\label{sec:action_labels}

Labels for longitudinal actions in DR(eye)VE and BDD-A were extracted automatically based on the provided speed data. We first removed outliers, applied a moving median filter with window of 20, and identified the points where speed changed using the method proposed in \cite{killick2012optimal}. We then computed acceleration between change points and identified braking and acceleration events using a threshold of 0.4 m/s\textsuperscript{2}. All episodes with acceleration below the threshold were labeled as \textit{maintain speed}. The \textit{stopped} label was assigned to the frames where speed did not exceed 1 km/h. Due to the limitations of the provided GPS and heading data, all lateral actions were labeled manually. 

Since LBW has no vehicle data,  we used the provided intrinsic and extrinsic camera matrices to infer camera movement via ORB-SLAM \cite{mur2015orb}. However, due to the missing frames and exposure issues (see Section \ref{sec:video_data}), only portions of the data could be processed. The resulting noisy speed and direction estimates were smoothed with a median filter but were of sufficient length to extract only the quick relative changes in speed over a few seconds, which likely inflated the proportion of \textit{maintain speed} episodes. The lateral actions (turns and lane changes) were labeled by hand.

Based on the summary of action statistics in Table \ref{tab:action_stats}, all three datasets are dominated by scenarios where the driver is \textit{maintaining speed/lane}, the vehicle is \textit{stopped}, or \textit{speeding up}, within its lane, usually after a stop. In DR(eye)VE, LBW, and BDD-A, these three actions comprise 81\%, 78\%, and 46\% of frames, respectively. Non-trivial scenarios, where drivers respond to traffic signals or interact with other road users, typically involve braking, turns, or lane changes. In BDD-A, only braking events are well-represented by design and comprise nearly half of the data. Lateral actions (turns and lane changes) in each dataset do not exceed 7\%. 

\begin{table}[]
\centering
\caption{Action statistics across the datasets. U-turns and reversing are excluded due to their rarity (\textless 1\% of the data). \textit{Turn/change lane}---actions performed while maintaining speed, \textit{lat/lon actions}---both lateral and longitudinal changes in motion are present (e.g. braking while making a turn).}
\label{tab:action_stats}
\begin{adjustbox}{max width=\columnwidth}
\begin{threeparttable}
\begin{tabular}{@{}lccc@{}}
\toprule
Action type         & \begin{tabular}[c]{@{}c@{}}DR(eye)VE / \\ MAAD\textsuperscript{a}\end{tabular} & BDD-A\textsuperscript{b} & LBW\textsuperscript{c}   \\ \midrule
Speed up            & 14.00                                                       & 14.05 & 9.63  \\
Slow down           & 12.46                                                       & 46.63 & 14.43 \\
Turn/change lane    & 2.99                                                        & 2.27  & 3.21  \\
Lateral/longitudinal actions     & 3.78                                                        & 4.66  & 3.66  \\
Maintain speed/lane & 60.59                                                       & 27.52 & 46.45 \\
Stopped             & 6.18                                                        & 4.86  & 22.62 \\ \bottomrule
\end{tabular}%
\begin{tablenotes}
\footnotesize
\item[a] Action counts are the same for MAAD since it uses DR(eye)VE videos. Video \# 6 was excluded due to mismatched vehicle data.
\item[b] Excluded 135 videos: the clips with severe image artifacts, missing gaze and vehicle data, and duplicates.
\item[c] Excluded two underexposed videos (Subject \# 26).
\end{tablenotes}
\end{threeparttable}
\end{adjustbox}
\vspace{-1.5em}
\end{table}

\subsection{Context}
We annotated the following context elements: intersections and the ego-vehicle priority when passing through them. For DR(eye)VE and BDD-A, we used the OpenStreetMap API \cite{OpenStreetMap} to extract the street network and identify intersections along the driven route for each video. We then visually inspected each intersection and assigned the following labels based on \cite{roess2014highway}: \textit{signalized} (if a traffic signal was present), \textit{unsignalized} (controlled by stop sign, yield sign, or no sign), \textit{roundabout}, and \textit{highway ramp}. We then marked the frame at which the vehicle passed through the middle of the intersection and ego-vehicle \textit{priority} (right-of-way or yielding) w.r.t. other vehicles based on the rules of the road in the recording location. Interactions with pedestrians were not considered because there were too few instances across all datasets. In LBW, context was annotated manually based on video recordings, since no GPS data was provided.

Table \ref{tab:context_stats} shows the summary statistics of intersections and the ego-vehicle priority in different datasets. With regard to context, the distribution across different types of intersections is largely location- and route-specific. For example, roundabouts are more common in Europe but not in North America, therefore in DR(eye)VE there are more such instances. In BDD-A, the videos are short and often truncated before the vehicles reaches an intersection, thus the number of intersections is comparatively low. Overall, the episodes where the vehicle is passing through some kind of intersection comprise about 7\% of the data in each dataset. In 70-80\% of the cases, the driver is traveling on the main road or passes through an intersection on a green traffic light and does not have to negotiate with other traffic participants.

\begin{table}[]
\centering
\caption{Number of instances of different types of intersections and the ego-vehicle right-of-way across the datasets.}
\label{tab:context_stats}
\begin{adjustbox}{max width=\columnwidth}
\begin{threeparttable}
\begin{tabular}{@{}llccc@{}}
\toprule
Intersection type & Priority & \begin{tabular}[c]{@{}c@{}}DR(eye)VE /\\ MAAD\end{tabular} & BDD-A & LBW \\ \midrule
\multirow{2}{*}{Unsignalized} & Right-of-way   & 436 & 211 & 463 \\
                              & Yield & 105 & 15  & 238 \\
\multirow{2}{*}{Signalized}   & Right-of-way   & 85  & 96  & 479 \\
                              & Yield & 13  & 124 & 53  \\
\multirow{2}{*}{Roundabout}   & Right-of-way   & 29  & -   & 5   \\
                              & Yield & 29  & -   & 5   \\
\multirow{2}{*}{Highway ramp} & Right-of-way   & 147 & -   & -   \\
                              & Yield & 12  & -   & -   \\ \bottomrule
\end{tabular}%
\begin{tablenotes}
\footnotesize
\item[*] Excluded videos are the same as in Table \ref{tab:action_stats}.
\end{tablenotes}
\end{threeparttable}
\end{adjustbox}
\vspace{-1.5em}
\end{table}

\section{Model performance}

\begin{table*}[t!]
\centering
\caption{Results of gaze prediction algorithms. $\uparrow$ and $\downarrow$ indicate that larger and smaller values are better. Best values are shown in \textbf{bold}. For subsets of data corresponding to different actions and context, cells are colored to highlight the \textcolor{ForestGreen}{best} and the \textcolor{BurntOrange}{worst} performance. Abbreviations: None---maintain speed/lane, Acc---longitudinal acceleration, Dec---longitudinal deceleration, Lat---lateral actions only, Lat/lon---simultaneous lateral and longitudinal action, Stop---ego-vehicle is stopped, RoW---ego-vehicle has right-of-way. Dashes (``--'') indicate that corresponding scenarios are not present in the test sets of BDD-A and LBW.}
\vspace{-1em}
\includegraphics[width=\textwidth]{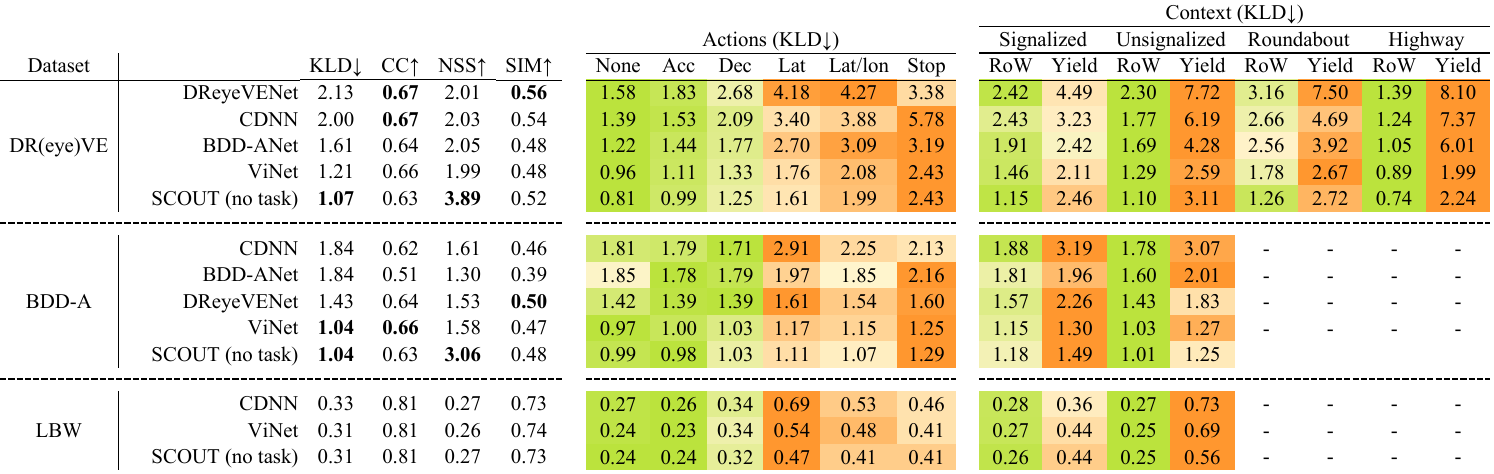}
\label{tab:benchmark_results}
\vspace{-1.5em}

\begin{flushleft}
\begin{footnotesize}
\textsuperscript{1}We use the corrected ground truth for DR(eye)VE and KLD implementation from \cite{fahimi2021metrics}. 

\textsuperscript{2} In all datasets, intersection scenarios (context) are gathered 1s prior to entering the intersection. Yielding scenarios in DR(eye)VE start from the first fixation on the intersecting road. Results on other metrics for actions and context show similar trends and will be released with the code for the paper. 

\textsuperscript{3}Official code for BDD-ANet and DReyeVENet could not be adapted to training on LBW.
\end{footnotesize}
\end{flushleft}
\vspace{-3em}
\end{table*}

\subsection{Experiment setup}
In this section, we use the newly proposed annotations to examine how well bottom-up models for drivers' gaze prediction capture the effects of task and context. We select a representative set of saliency models with publicly available code: DReyeVENet \cite{2018_PAMI_Palazzi}, BDD-ANet \cite{2018_ACCV_Xia}, ViNet \cite{2021_IROS_Jain}, CDNN \cite{2016_T-ITS_Deng}, and SCOUT \cite{kotseruba2023understanding}. Evaluation is done using common saliency metrics: Kullback--Leibler divergence (KLD), Pearson's correlation coefficient (CC), histogram similarity (SIM), and Normalized Scanpath Saliency (NSS) \cite{2018_PAMI_Bylinskii}. All models are trained on each dataset using the official code, train/test splits, and default hyperparameters. Input is either a single frame (for CDNN) or a stack of frames representing 0.5s observation (for DReyeVENet, BDD-ANet, ViNet, and SCOUT)\footnote{Since LBW has no official train/test split, we randomly selected videos of subjects 15, 19, 20, 22, and 25 for testing, 16 and 17 for validation, and the rest for training.}. Since LBW data is subsampled to 5Hz, 3 frames correspond to 0.5s observation. To accommodate the fixed input size of ViNet and SCOUT, the first 13 frames of input are filled with zeros. Prediction is made for a single input frame or the last frame in a stack. All frames containing U-turns and reversing are excluded from training and evaluation due to the rarity of such events. 

\subsection{Evaluation results}
\noindent
\textbf{The overall performance is high, particularly on LBW, due to specifics of its ground truth.} Results in Table \ref{tab:benchmark_results} demonstrate that the overall performance of the models is comparable on DR(eye)VE and BDD-A but very different on LBW. On the latter, the three tested models perform much more uniformly and achieve much better results on distribution-based metrics KLD, CC, and SIM. However, location-based NSS is reduced almost by one order of magnitude for some models. The most likely explanation is the nature of the LBW saliency maps, which are based on a single fixation and cover a large part of the image (see Figure \ref{fig:salmap_examples}). Because of this, even minor mismatches are heavily penalized by NSS, since it is sensitive to false positives. At the same time, the broad extent of the saliency maps is beneficial for the distribution-based metrics that measure intersection (SIM), correlation (CC), or difference (KLD) between ground truth and prediction. In other words, the models do not need to be precise to achieve good results.

\noindent
\textbf{Performance drops on scenarios with lateral actions and yielding.} All models across all three datasets perform better on action categories \textit{maintain}, \textit{accelerate} and \textit{decelerate} than on sequences with lateral actions and stops. Similarly, performance is significantly worse on yielding scenarios compared to right-of-way ones. Although it is tempting to attribute these differences to the distribution of training data, the correlation between the results and data composition in Tables \ref{tab:action_stats} and \ref{tab:context_stats} is not statistically significant.

Complex interactions with other road users are likely not a factor for the poor performance of yielding and lateral actions. We estimated that only 27\%, 24\% and 22\% of such scenarios in DR(eye)VE, LBW, and BDD-A involve at least one conflict vehicle, the rest are on empty roads.

\noindent
\textbf{Data limitations are likely the key factor affecting the results.} We argue that the limited spatial context and loss of eye-tracking data discussed in Section \ref{sec:data_analysis} contribute the most to the observed performance trends. To recap, near intersections, drivers often look at areas and objects that are not visible in the scene camera view. Some drivers' gaze information can be preserved, for example, in DR(eye)VE, fixations that fall outside the camera view were pushed towards the edges of the frame to indicate direction and elevation of gaze \cite{kotseruba2023understanding}. However, there are no cues in the images to associate gaze with. 

When videos from the DR(eye)VE dataset were shown to observers in the lab \cite{2021_ICCVW_Gopinath}, this lack of context translated to different gaze patterns (see Section \ref{sec:data_recording} and Figures \ref{fig:road_lab},\ref{fig:road_lab_sample}). When yielding at the intersections or changing lanes, many subjects did not check the intersecting road since only a small part of it was visible on the monitor and did not see conflicting vehicles until they appeared, thus their gaze was more centered. We speculate that the same is true for BDD-A, which is recorded in the lab and where many scenarios occur near intersections. This may explain why the gap between right-of-way and yielding sequences and lateral and longitudinal actions in BDD-A is smaller than in DR(eye)VE. While the gap between lateral and longitudinal actions in LBW is also small, in this dataset there is significant data loss around lateral maneuvers and major differences in how the saliency maps are computed (see Sections \ref{sec:video_data} and \ref{sec:data_processing}), which prevent direct comparisons. Further investigation of these issues is warranted.

There is also evidence that neither use of additional features nor weighting techniques compensate for these data limitations. For instance, DReyeVENet relies on optical flow and semantic segmentation maps for scene analysis and BDD-ANet implements sample weighting to penalize errors on frames with ``unusual'' gaze patterns. However, both models are outperformed by ViNet and SCOUT, which use only images and a similar spatio-temporal feature extraction.

\section{Discussion and Future Work}
In this paper, we analyzed four publicly available datasets for drivers' gaze prediction and evaluated the ability of models trained on this data to capture the effects of the underlying task performed by the driver. We found a number of limitations in the data collection and processing pipelines. In addition, the data itself is dominated by trivial scenarios, where driver maintains speed or lane, or the vehicle is stopped. We then linked reduced performance of all tested models on scenarios involving lateral actions and intersections to these data limitations. Based on our findings, we argue that simply increasing the proportion of non-trivial scenarios in data will likely not suffice unless other issues with the data recording, processing, and annotation are resolved first. These conclusions apply not only to gaze prediction, but also to object importance estimation and explainability, since they are likewise related to drivers' actions/attention and use similar data for training and evaluation. 

%

Based on the identified issues, we propose the following key desiderata for data collection and processing: 1) temporal and spatial context should be sufficient, \ie videos should be several minutes long and record 180$^\circ$ view around the vehicle with multiple synchronized cameras, 2) vehicle telemetry should include accurate GPS data, speed, IMU sensor output, brake/throttle pedal, steering wheel rotations, and turn signals to better represent drivers' actions, 3) on-road gaze recordings must be accurately matched to the scene view using camera calibration information and labeled for episodes of distractions and glances to the car interior, 4) in-lab gaze recordings should be accompanied by the same instructions that the drivers on the road received, and, lastly, 5) raw data should be made available for re-analysis and reproducibility.





\noindent
\textbf{Acknowledgement.} Work supported by the Air Force Office of Scientific Research under award number FA9550-22-1-0538 (Computational Cognition and Machine Intelligence, and Cognitive and Computational Neuroscience portfolios); the Canada Research Chairs Program (950-231659); Natural Sciences and Engineering Research Council of Canada (RGPIN-2022-04606).

\bibliographystyle{IEEEtran}
\bibliography{references}

\begin{thebibliography}{10}
\providecommand{\url}[1]{#1}
\csname url@samestyle\endcsname
\providecommand{\newblock}{\relax}
\providecommand{\bibinfo}[2]{#2}
\providecommand{\BIBentrySTDinterwordspacing}{\spaceskip=0pt\relax}
\providecommand{\BIBentryALTinterwordstretchfactor}{4}
\providecommand{\BIBentryALTinterwordspacing}{\spaceskip=\fontdimen2\font plus
\BIBentryALTinterwordstretchfactor\fontdimen3\font minus
  \fontdimen4\font\relax}
\providecommand{\BIBforeignlanguage}[2]{{%
\expandafter\ifx\csname l@#1\endcsname\relax
\typeout{** WARNING: IEEEtran.bst: No hyphenation pattern has been}%
\typeout{** loaded for the language `#1'. Using the pattern for}%
\typeout{** the default language instead.}%
\else
\language=\csname l@#1\endcsname
\fi
#2}}
\providecommand{\BIBdecl}{\relax}
\BIBdecl

\bibitem{sivak1996information}
M.~Sivak, ``The information that drivers use: is it indeed 90\% visual?''
  \emph{Perception}, vol.~25, no.~9, pp. 1081--1089, 1996.

\bibitem{cunningham2001driving}
D.~W. Cunningham, A.~Chatziastros, M.~Von~der Heyde, and H.~H. B{\"u}lthoff,
  ``Driving in the future: temporal visuomotor adaptation and generalization,''
  \emph{Journal of Vision}, vol.~1, no.~2, pp. 88--98, 2001.

\bibitem{lappi2017systematic}
O.~Lappi, P.~Rinkkala, and J.~Pekkanen, ``Systematic observation of an expert
  driver's gaze strategy --- an on-road case study,'' \emph{Frontiers in
  Psychology}, vol.~8, p. 620, 2017.

\bibitem{lappi2018visuomotor}
O.~Lappi and C.~Mole, ``Visuomotor control, eye movements, and steering: A
  unified approach for incorporating feedback, feedforward, and internal
  models,'' \emph{Psychological Bulletin}, vol. 144, no.~10, pp. 981--1001,
  2018.

\bibitem{shinoda2001controls}
H.~Shinoda, M.~M. Hayhoe, and A.~Shrivastava, ``What controls attention in
  natural environments?'' \emph{Vision Research}, vol.~41, no. 25-26, pp.
  3535--3545, 2001.

\bibitem{baluch2011mechanisms}
F.~Baluch and L.~Itti, ``Mechanisms of top-down attention,'' \emph{Trends in
  Neurosciences}, vol.~34, no.~4, pp. 210--224, 2011.

\bibitem{metz2017driving}
B.~Metz, N.~Schoemig, and H.-P. Krueger, ``How is driving-related attention in
  driving with visual secondary tasks controlled? {E}vidence for top-down
  attentional control,'' in \emph{Driver distraction and inattention}.\hskip
  1em plus 0.5em minus 0.4em\relax CRC Press, 2017, pp. 83--102.

\bibitem{kotseruba2022practical}
I.~Kotseruba and J.~K. Tsotsos, ``Attention for vision-based assistive and
  automated driving: A review of algorithms and datasets,'' \emph{IEEE T-ITS},
  vol.~23, no.~11, pp. 19\,907--19\,928, 2022.

\bibitem{ning2019efficient}
M.~Ning, C.~Lu, and J.~Gong, ``An efficient model for driving focus of
  attention prediction using deep learning,'' in \emph{ITSC}, 2019.

\bibitem{2020_T-ITS_Deng}
T.~Deng, H.~Yan, L.~Qin, T.~Ngo, and B.~Manjunath, ``How do drivers allocate
  their potential attention? {D}riving fixation prediction via convolutional
  neural networks,'' \emph{IEEE T-ITS}, vol.~21, no.~5, pp. 2146--2154, 2019.

\bibitem{2018_PAMI_Palazzi}
A.~Palazzi, D.~Abati, F.~Solera, R.~Cucchiara \emph{et~al.}, ``{Predicting the
  Driver's Focus of Attention: the DR (eye) VE Project},'' \emph{IEEE TPAMI},
  vol.~41, no.~7, pp. 1720--1733, 2018.

\bibitem{2021_T-ITS_Fang}
J.~Fang, D.~Yan, J.~Qiao, J.~Xue, and H.~Yu, ``{DADA}: {D}river attention
  prediction in driving accident scenarios,'' \emph{IEEE T-ITS}, vol.~23, pp.
  4959--4971, 2022.

\bibitem{2021_T-ITS_Amadori}
P.~V. Amadori, T.~Fischer, and Y.~Demiris, ``Hammerdrive: A task-aware driving
  visual attention model,'' \emph{IEEE T-ITS}, vol.~23, pp. 5573--5585, 2021.

\bibitem{2021_ICCV_Baee}
S.~Baee, E.~Pakdamanian, I.~Kim, L.~Feng, V.~Ordonez, and L.~Barnes,
  ``{MEDIRL}: {P}redicting the visual attention of drivers via maximum entropy
  deep inverse reinforcement learning,'' in \emph{ICCV}, 2021.

\bibitem{kotseruba2023understanding}
I.~Kotseruba and J.~K. Tsotsos, ``Understanding and modeling the effects of
  task and context on drivers' gaze allocation,'' in \emph{IV}, 2024.

\bibitem{2014_JEMR_Lemonnier}
S.~Lemonnier, R.~Br{\'e}mond, and T.~Baccino, ``Discriminating cognitive
  processes with eye movements in a decision-making driving task,''
  \emph{Journal of Eye Movement Research}, vol.~7, no.~4, pp. 1--14, 2014.

\bibitem{2015_TR_Lemonnier}
S.~Lemonnier, R.~Br{\'e}mond, and T.~Baccino, ``{Gaze behavior when approaching
  an intersection: Dwell time distribution and comparison with a quantitative
  prediction},'' \emph{Trans. Res. Part F: Traffic Psychology and Behaviour},
  vol.~35, pp. 60--74, 2015.

\bibitem{attention_driving_datasets}
I.~Kotseruba, ``Attention and driving,''
  \url{https://github.com/ykotseruba/attention_and_driving/blob/2.0/datasets.md},
  2022.

\bibitem{2018_ACCV_Xia}
Y.~Xia, D.~Zhang, J.~Kim, K.~Nakayama, K.~Zipser, and D.~Whitney, ``Predicting
  driver attention in critical situations,'' in \emph{ACCV}, 2018.

\bibitem{2021_ICCVW_Gopinath}
D.~Gopinath, G.~Rosman, S.~Stent, K.~Terahata, L.~Fletcher, B.~Argall, and
  J.~Leonard, ``{MAAD: A model and dataset for "attended awareness" in
  driving},'' in \emph{ICCVW}, 2021.

\bibitem{2022_ECCV_Kasahara}
I.~Kasahara, S.~Stent, and H.~S. Park, ``{Look Both Ways: Self-supervising
  driver gaze estimation and road scene saliency},'' in \emph{ECCV}, 2022.

\bibitem{xu2017end}
H.~Xu, Y.~Gao, F.~Yu, and T.~Darrell, ``End-to-end learning of driving models
  from large-scale video datasets,'' in \emph{CVPR}, 2017.

\bibitem{2018_ITSC_Tawari}
A.~Tawari, P.~Mallela, and S.~Martin, ``Learning to attend to salient targets
  in driving videos using fully convolutional {RNN},'' in \emph{ITSC}, 2018.

\bibitem{2020_iPerception_Strasburger}
H.~Strasburger, ``Seven myths on crowding and peripheral vision,''
  \emph{i-Perception}, vol.~11, no.~3, p. 2041669520913052, 2020.

\bibitem{kotseruba2021behavioral}
I.~Kotseruba and J.~K. Tsotsos, ``Behavioral research and practical models of
  drivers' attention,'' \emph{arXiv:2104.05677}, 2021.

\bibitem{ho2014extent}
C.~Ho, R.~Gray, and C.~Spence, ``To what extent do the findings of
  laboratory-based spatial attention research apply to the real-world setting
  of driving?'' \emph{IEEE Trans. HMS}, vol.~44, no.~4, pp. 524--530, 2014.

\bibitem{mit-saliency-benchmark}
Z.~Bylinskii, T.~Judd, A.~Borji, L.~Itti, F.~Durand, A.~Oliva, and A.~Torralba,
  ``{MIT Saliency Benchmark}.''

\bibitem{mathe2014actions}
S.~Mathe and C.~Sminchisescu, ``Actions in the eye: Dynamic gaze datasets and
  learnt saliency models for visual recognition,'' \emph{IEEE TPAMI}, vol.~37,
  no.~7, pp. 1408--1424, 2014.

\bibitem{xu2014predicting}
J.~Xu, M.~Jiang, S.~Wang, M.~S. Kankanhalli, and Q.~Zhao, ``Predicting human
  gaze beyond pixels,'' \emph{Journal of Vision}, vol.~14, no.~1, pp. 28--28,
  2014.

\bibitem{nguyen2019saliency}
A.~Nguyen and Z.~Yan, ``A saliency dataset for 360-degree videos,'' in
  \emph{ACM Multimedia Systems Conference}, 2019, pp. 279--284.

\bibitem{killick2012optimal}
R.~Killick, P.~Fearnhead, and I.~A. Eckley, ``Optimal detection of changepoints
  with a linear computational cost,'' \emph{Journal of the American Statistical
  Association}, vol. 107, no. 500, pp. 1590--1598, 2012.

\bibitem{mur2015orb}
R.~Mur-Artal, J.~M.~M. Montiel, and J.~D. Tardos, ``{ORB-SLAM}: {A} versatile
  and accurate monocular slam system,'' \emph{IEEE Transactions on Robotics},
  vol.~31, no.~5, pp. 1147--1163, 2015.

\bibitem{OpenStreetMap}
{OpenStreetMap contributors}, ``{Planet dump retrieved from
  https://planet.osm.org },'' \url{ https://www.openstreetmap.org }, 2017.

\bibitem{roess2014highway}
R.~P. Roess and E.~S. Prassas, ``The highway capacity manual: {A} conceptual
  and research history,'' 2014.

\bibitem{fahimi2021metrics}
R.~Fahimi and N.~D. Bruce, ``On metrics for measuring scanpath similarity,''
  \emph{Behav. Res. Methods}, vol.~53, no.~2, pp. 609--628, 2021.

\bibitem{2021_IROS_Jain}
S.~Jain, P.~Yarlagadda, S.~Jyoti, S.~Karthik, R.~Subramanian, and V.~Gandhi,
  ``{ViNet: Pushing the limits of visual modality for audio-visual saliency
  prediction},'' in \emph{IROS}, 2021.

\bibitem{2016_T-ITS_Deng}
T.~Deng, K.~Yang, Y.~Li, and H.~Yan, ``Where does the driver look?
  top-down-based saliency detection in a traffic driving environment,''
  \emph{IEEE T-ITS}, vol.~17, no.~7, pp. 2051--2062, 2016.

\bibitem{2018_PAMI_Bylinskii}
Z.~Bylinskii, T.~Judd, A.~Oliva, A.~Torralba, and F.~Durand, ``What do
  different evaluation metrics tell us about saliency models?'' \emph{IEEE
  TPAMI}, vol.~41, no.~3, pp. 740--757, 2018.

\end{thebibliography}

\end{document}